\documentclass[conference]{IEEEtran}
\IEEEoverridecommandlockouts
\usepackage{cite}
\usepackage{amsmath,amssymb,amsfonts}
\usepackage{algorithmic}
\usepackage{graphicx}
\usepackage{textcomp}
\usepackage{xcolor}
\usepackage[ruled,vlined]{algorithm2e}
\usepackage{multirow,makecell}
\usepackage{booktabs}
\usepackage{comment}

\newcommand{\hw}[1]{\ensuremath{\mathtt{#1}}}

\def\BibTeX{{\rm B\kern-.05em{\sc i\kern-.025em b}\kern-.08em
    T\kern-.1667em\lower.7ex\hbox{E}\kern-.125emX}}
\begin{document}

\title{
Mix and Match: A Novel FPGA-Centric Deep Neural Network Quantization Framework
}

\author{\IEEEauthorblockN{Sung-En Chang$^{*1}$, Yanyu Li$^{*1}$, Mengshu Sun$^{*1}$, \thanks{$^*$Equal contribution. }Runbin Shi$^2$, Hayden K.-H. So$^2$, Xuehai Qian$^3$,\\ Yanzhi Wang$^1$,
Xue Lin$^1$}
\IEEEauthorblockA{\textit{$^1$Northeastern University, $^2$The University of Hong Kong, $^3$University of Southern California} \\
\{chang.sun, li.yanyu, sun.meng, yanz.wang, xue.lin\}@northeastern.edu, \{rbshi, hso\}@eee.hku.hk, xuehai.qian@usc.edu}
}

\maketitle

\begin{abstract}
Deep Neural Networks (DNNs) have achieved extraordinary performance in various application domains.
To support diverse DNN models, 
efficient implementations of DNN inference on edge-computing platforms, e.g., ASICs, FPGAs, and embedded systems, are extensively investigated. 
Due to the huge model size and computation amount,
model compression is a critical step to deploy 
DNN models on edge devices.
This paper focuses on weight quantization, a hardware-friendly model compression approach that is complementary to weight pruning.

Unlike existing methods that use the same quantization
scheme for all weights, we propose the {\em first} solution
that applies different quantization schemes for 
different rows of the weight matrix. 
It is motivated by (1) the distribution of the weights 
in the different rows are not the same; and
(2) the potential of achieving better utilization of
heterogeneous FPGA hardware resources. 
To achieve that, we first propose a hardware-friendly quantization scheme named {\em sum-of-power-of-2 (SP2)}
suitable for Gaussian-like weight distribution, in which the
multiplication arithmetic can be replaced with logic shifter and adder, thereby enabling highly efficient implementations with the FPGA LUT resources.
In contrast, the existing fixed-point quantization is suitable
for Uniform-like weight distribution and can be implemented
efficiently by DSP.
Then to fully explore the resources, we propose an FPGA-centric mixed scheme quantization (MSQ) with an ensemble of the proposed SP2 and the fixed-point schemes. 
Combining the two schemes can maintain, or even increase accuracy due to better matching with 
weight distributions. 

For the FPGA implementations, we develop a parameterized architecture with heterogeneous Generalized Matrix Multiplication (GEMM) cores---one using LUTs for computations with SP2 quantized weights and the other utilizing DSPs for fixed-point quantized weights. 
Given the partition ratio among the two schemes based on 
resource characterization, MSQ quantization training algorithm derives an optimally quantized model for the FPGA implementation.
We evaluate our FPGA-centric quantization framework across multiple application domains.  
With optimal SP2/fixed-point ratios on two FPGA devices, i.e., Zynq XC7Z020 and XC7Z045, we achieve performance improvement of $\mathbf{2.1\times-4.1\times}$ compared to solely exploiting DSPs for all multiplication operations. In addition, the CNN implementations with the proposed MSQ scheme can achieve higher accuracy and comparable hardware utilization efficiency compared to the state-of-the-art designs.
\end{abstract}


\section{Introduction}

{Deep learning or Deep Neural Networks (DNNs) have achieved extraordinary performance in various application domains \cite{szegedy2017inception,lin2017feature,hinton2012deep,xiong2018microsoft,collobert2008unified,litjens2017survey,de2018clinically}.
However, the state-of-the-art DNNs may require up to GBs (Giga Bytes) for model size and $10^2$ GFLOPs (Giga Floating Point Operations) for inference computation, making it a
challenging task to perform on-device inference.}

{To efficiently execute the diverse DNN inference models
for broader applications, 
the resource-constrained edge computing platforms require
two crucial supports.
The first one is the specialized hardware acceleration for DNN
inference. 
Extensive research efforts have been dedicated to the efficient implementations of DNN inference models on various edge-computing platforms, such as ASICs
~\cite{mao2018lergan,ren2017sc,cai2018vibnn,sharma2018bit,deng2019tie,cai2019stochastic,ren2019admm},
FPGAs
~\cite{zhang2015optimizing,zhao2017accelerating,chen2018fpga,shi2020ftdl},
and embedded CPUs/GPUs
~\cite{niu2020patdnn,chen2018tvm,Ali-MNN,paszke2017pytorch,TensorFlow-Lite}.
}

The second is the DNN model compression technique, 
which not only seeks more efficient hardware implementation
based on given models, but also explores the opportunity
of algorithm and hardware co-design to achieve 
better trade-offs among accuracy, hardware cost, and performance.
There are two essential techniques for model compression:
DNN weight pruning~{\cite{liu2018rethinking,wen2016learning,guo2016dynamic,zhuang2018discrimination,yu2018nisp,he2019filter,dong2019network}} and weight quantization \cite{courbariaux2015binaryconnect,courbariaux2016binarized,rastegari2016xnor,lin2017towards,li2016ternary,zhu2016trained,he2019simultaneously,zhou2016dorefa,choi2018pact,gong2019differentiable,jung2019learning,cheng2019uL2Q,esser2019learned,zhang2018lq,DBLP:journals/corr/MiyashitaLM16,DBLP:journals/corr/ZhouYGXC17,leng2018extremely}.

This paper focuses on DNN weight quantization, which becomes imperative to the DNN hardware acceleration especially on the FPGA and ASIC platforms.
By representing weights with fewer bits, 
weight quantization can directly simplify the implementations and accelerate the inference execution speed in a 
{\em hardware-friendly manner}. Also, it is supported in GPUs (e.g., PyTorch \cite{paszke2017pytorch} for NVIDIA GPUs) and mobile devices (e.g., TensorFlow-Lite \cite{TensorFlow-Lite}).
In addition, weight quantization yields far less training overhead than weight pruning, let alone the training-heavy network architecture search (NAS)-based model compression techniques. Specifically, in state-of-the-art DNN quantization methods (including our work), retraining process takes usually $1/3\sim1/2$ of the epochs as those for the pre-training process, which is totally acceptable training overhead in the exchange for significant inference speedup.

{Weight quantization can be considered as a mapping from 32-bit floating-point weights 
into $m$-bit weight representations.
There are different types of quantization schemes including binary \cite{courbariaux2015binaryconnect,courbariaux2016binarized,rastegari2016xnor,lin2017towards}, ternary \cite{li2016ternary,zhu2016trained,he2019simultaneously}, low-bit-width fixed-point \cite{zhou2016dorefa,choi2018pact,gong2019differentiable,jung2019learning,cheng2019uL2Q,esser2019learned}, and power-of-2 \cite{zhang2018lq,DBLP:journals/corr/MiyashitaLM16,DBLP:journals/corr/ZhouYGXC17,leng2018extremely}.
In general, binary and ternary quantization schemes result in significant accuracy loss, for example, $>5\%$ under binary and $2\%-3\%$ for ternary quantization.
The fixed-point quantization can represent the DNN weights using low bit-width, e.g., 4-bit, with negligible accuracy loss.
To further simplify hardware implementations, power-of-2 quantization scheme was proposed to replace the multiplications with bit-shifting operations. However, power-of-2 results in non-negligible accuracy degradation, usually around $1\%-2\%$, which even cannot be overcome with increasing precision.
}

To overcome the challenges, instead of using the same quantization
scheme for all weights, we propose the {\em first} solution
that applies different quantization 
schemes for 
different rows of the weight matrix. 
It is motivated by (1) the distribution of weights 
in the different rows are not the same; and 
(2) the potential of achieving better utilization of 
heterogeneous FPGA hardware resources. 
We propose a hardware-friendly quantization scheme named {\em sum-of-power-of-2 (SP2)}
suitable for Gaussian-like weight distribution, in which the
multiplication arithmetic can be replaced with logic shifter and adder, thereby enabling highly efficient implementations with the FPGA LUT resources.
At the same time, the SP2 quantization enjoys the negligible accuracy loss, just like the fixed-point quantization scheme.
In comparison, the fixed-point quantization is suitable
for Uniform-like weight distribution and can be implemented
efficiently by DSP.

To fully explore the FPGA resources, we propose an FPGA-centric mixed scheme quantization (MSQ) with an ensemble of the proposed SP2 and the fixed-point schemes. 
Given that each individual scheme can achieve negligible accuracy loss,
we demonstrate that combining the two can maintain, or 
even reach higher accuracy.
It is due to the benefit of using two quantization schemes:
even within a single layer, the local weight distributions can be diverse, if we assign the right quantization scheme 
to better fit the local weight distributions, accuracy can be boosted.

For the FPGA implementations, we developed a parameterized architecture with heterogeneous Generalized Matrix Multiplication (GEMM) cores---one using LUTs for computations with SP2 quantized weights and the other utilizing DSPs for fixed-point quantized weights. 
We first find the partition ratio of the SP2 to fixed-point quantization for weights of a DNN layer through FPGA resource characterization, such that the DSP utilization is kept at 100\% and LUT utilization can also be optimized. 
Given the partition ratio, MSQ quantization training algorithm derives an optimally quantized model for the FPGA implementation. We evaluate our FPGA-centric quantization framework across multiple application domains including image classification, object detection and recognition, machine translation, speech recognition, sentiment classification, and natural language processing, with various DNNs such as convolutional neural networks (CNN), and recurrent neural networks (RNN). 
With optimal SP2/fixed-point ratios on two FPGA devices, i.e., Zynq XC7Z020 and XC7Z045, we achieve performance improvement of $\mathbf{2.1\times-4.1\times}$ compared to solely exploiting DSPs for all multiplication operations. In addition, the CNN implementations with the proposed MSQ scheme can achieve higher accuracy and comparable hardware utilization efficiency compared to state-of-the-arts.

The contributions of this work are:
\begin{itemize}
    \item We propose a novel hardware-friendly SP2 quantization scheme, which enjoys both non-multiplication operations and negligible accuracy degradation.
    \item We provide the first DNN quantization solution that jointly applies two quantization schemes to achieve better utilization of heterogeneous FPGA hardware resources while not harming the quantized model accuracy.
    \item Our framework features a novel architecture with heterogeneous GEMM engines and design optimizations, to accommodate our mixed scheme quantization and to optimize FPGA resource allocation.
    \item The effectiveness of our proposed MSQ is validated across multiple application domains and with FPGA devices, on inference accuracy and FPGA resource utilization efficiency.
\end{itemize}

Our work is significantly different from existing quantization frameworks that leverage the inter-layer, multi-precision approach. We exploit the previously neglected flexibility on quantization schemes (using both fixed-point and SP2) by adopting a novel \emph{intra-layer, multi-scheme} approach. Specifically, we identify an optimized ratio of the two schemes from FPGA (LUT and DSP) resource characterization, and then assign the different rows of the weight matrix \emph{within a layer} into the two schemes according to the weight distributions. Our method is totally perpendicular to, and can be combined with, the existing inter-layer, multi-precision approaches.

\section{Background on DNN Weight Quantization}\label{sec:background}

\subsection{Weight Quantization Schemes}

\subsubsection{Uniform Interval Quantization Schemes}\label{sec:uniformschemes}

{Uniform interval quantization schemes include binary, ternary, and low-bit-width fixed-point.
Binary or ternary quantization uses extremely low precision for DNN models, i.e., binarized (e.g., -1, +1) or ternarized (e.g., -1, 0, +1) levels.
Representative binary quantization methods include Binaryconnect \cite{courbariaux2015binaryconnect}, Binarized Neural Network (BNN) \cite{courbariaux2016binarized}, XNOR-net \cite{rastegari2016xnor}, and ABC-Net \cite{lin2017towards}.
With weights constrained to $\{-1, 1\}$, multiplications can be replaced by additions/subtractions. Additions/subtractions can also be eliminated using XNOR and AND operations if activations are quantized to binary as well. 
On the other hand, ternary quantization schemes are implemented in TWN \cite{li2016ternary}, TTQ \cite{zhu2016trained}, and  \cite{he2019simultaneously}.
Ternary representation keeps zero in quantization levels, which requires one more bit to present weights. Ternary networks also benefit from non-multiplication operations 
while maintaining the natural sparsity (since zero weights are kept). Although binary and ternary quantization can significantly reduce operations and simplify the implementations of hardware accelerators, it introduces non-negligible accuracy loss. For example, based on reports from the above works, accuracy typically degrades by $>5\%$ under the binary scheme, and $2-3\%$ for ternary.
}

{Comparing with binary and ternary quantization, the fixed-point quantization scheme applies the modest and flexible quantization rates to preserve the accuracy as that of the 32-bit floating-point models.
For example, 4-bit fixed-point introduces zero or negligible accuracy loss.
Fixed-point quantization scheme has been implemented with different methods/algorithms by DoReFa-Net \cite{zhou2016dorefa}, PACT \cite{choi2018pact}, DSQ \cite{gong2019differentiable}, QIL \cite{jung2019learning}, $\mu$ L2Q \cite{cheng2019uL2Q}, and LSQ \cite{esser2019learned}.

With the $m$-bit fixed-point scheme, quantized weight values are defined as the scaling factor $\alpha$ times quantization levels: 
\begin{equation}\label{eq:fixedpointQL}
\mathcal{Q}^{FP}(m, \alpha)= \pm\alpha \times \{0, \frac{1}{2^{m-1}-1}, \frac{2}{2^{m-1}-1}, ...,  1\}.
\end{equation}
And the mapping from a 32-bit floating-point weight $w$ into the quantized weight $\hat w$ by $m$-bit fixed-point representation (in sign-magnitude) is given by the following quantizer:
\begin{equation}\label{eq:fixedpointquantizer}
\begin{aligned}
\hat w  &= \prod_{\mathcal{Q}^{FP}(m, \alpha)} w\\
 &=\alpha\cdot h^{-1}\big(\frac{1}{2^{m}-1} round((2^{m}-1) \cdot h(\lceil w,\alpha\rfloor))\big),
 \end{aligned}
\end{equation}
where $\prod_{\mathcal{Q}^{FP}(m,\alpha)}(\cdot)$ denotes the quantizer function to project onto $\mathcal{Q}^{FP}(m,\alpha)$; the function $h(\cdot)$ transforms a value within $[-1,+1]$ into the range of $[0,1]$, for example we can use $h(\cdot)=\text{tanh}(\cdot)/2+0.5$; and $\lceil w,\alpha\rfloor$ clips $w$ according to
\begin{equation}
    \begin{aligned}
    \lceil w,\alpha\rfloor = 
    \begin{cases}
    -1, \quad &w<-\alpha\\
    w/\alpha, \quad &-\alpha\leq w \leq\alpha\\
    1, \quad & w>\alpha
    \end{cases}
    \end{aligned}
.\end{equation}

}

\subsubsection{Non-Uniform Interval Quantization Schemes}

{On the other hand, power-of-2 quantization is a non-uniform interval quantization scheme, representative methods including \cite{zhang2018lq,DBLP:journals/corr/MiyashitaLM16,DBLP:journals/corr/ZhouYGXC17,leng2018extremely}.
Power-of-2 quantization replaces multiplications by bit shifting operations and this number system also possesses higher precision around the mean, which fits the Gaussian distribution of DNN weights better \cite{baskin2018uniq,blundell2015weight}.
With an $m$-bit weight representation (in sign-magnitude), the quantized weight values by the power-of-2 scheme are defined as
\begin{equation}\label{eq:P2QL}
Q^{P2}(m, \alpha)= \pm\alpha\times \{0, \frac{ 1}{2^{2^{m-1}-2}}, \frac{ 1}{2^{2^{m-1}-3}}, ...,  1\}
.\end{equation}
And the power-of-2 quantizer is then given by

\begin{equation}\label{eq:powerof2quantizer}
\begin{aligned}
\hat w
&= \prod_{\mathcal{Q}^{P2}(m, \alpha)} w\\
&=\begin{cases}
\begin{aligned}
\alpha\cdot h^{-1}\big(2^{round(\log_2 h(\lceil w,\alpha\rfloor))}\big) \quad &h(\lceil w,\alpha\rfloor)>2^{-2^m+1}\\
0 \quad &h(\lceil w,\alpha\rfloor)\leq 2^{-2^m+1}\\
\end{aligned}.
\end{cases}
\end{aligned}
\end{equation}
}

{With weights quantized into the power-of-2 scheme, multiplications between weight i.e., $2^b  (b\in \mathbb{N})$ and activation i.e., $a$ can be implemented by bit shifting as follows:
\begin{equation}
\begin{aligned}
2^b\times a=\begin{cases}
a<<b, \quad &b>0\\
a, \quad &b=0\\
a>>b, \quad &b<0\\
\end{cases}
\end{aligned}
.\end{equation}
Although the power-of-2 quantization scheme can simplify hardware implementation by eliminating multiplications, its precision cannot be increased effectively with increasing $m$, because increasing $m$ will merely increase resolution around the mean, while the tails are still in low precision. This can also be observed from Eq (\ref{eq:powerof2quantizer}) that when $w$ is a large value, increasing $m$ does not have an effect on $\hat w$.
In practice, $3\sim7$ bits are usually used for power-of-2 quantization, and more bits could not further promote the accuracy of the quantized models.
{As mentioned in \S\ref{sec:uniformschemes} that 4-bit fixed-point results in negligible accuracy degradation, but 4-bit power-of-2 quantization will result in accuracy loss of $1\%-2\%$.}
}

\subsection{Quantization Algorithms}

\begin{algorithm}[t]
\small
\caption{\small DNN Quantization with ADMM and STE}\label{algo:ADMMQuantization}
\SetKwInOut{Input}{input}
\SetKwInOut{Output}{target}
\SetKwFunction{SGD}{SGD}
\Input{
\noindent 32-bit floating-point DNN model $\mathcal{M}$, with weights  $\mathbf{W}$ to be quantized.\\
\noindent Quantization scheme: $\mathbb{S}\in\{$Fixed-point, Power-of-2, Sum-of-power-of-2$\}$\\
}
\Output{ Quantized model $\hat{\mathcal{M}}$ }
\BlankLine

\tcp{Initialization:} 
$U^0=0$; $Z^0=\mathbf{W}$;
\BlankLine
\ForEach{Epoch}{
\tcp{Update Z, U:}
$Z^{t}\leftarrow$ $\textbf{proj}_\mathbb{S}$($\mathbf{W}+U^{t-1}$);\\
$U^{t}\leftarrow \mathbf{W}-Z^{t}+U^{t-1}$;\\
\ForEach{Batch}{
\tcp{STE for activation quantization:}
$input \leftarrow \textbf{proj}_\mathbb{S}(input)$;\\ 
$loss\leftarrow \mathcal{M}(input)$;\\
$loss\leftarrow loss+\sum \frac{1}{2} \rVert \mathbf{W}-Z^t+U^t \rVert^2$;\\
Backpropagate $loss$ and update $\mathbf{W}$;\\
\BlankLine
}
}

Return $\hat{\mathcal{M}} \leftarrow \mathcal{M}\{\textbf{proj}_\mathbb{S}(\mathbf{W})\}$.

\end{algorithm}

{Quantization performs projection from the continuous domain to a discrete number system, which makes the gradients of the loss function unavailable for backpropagation during the training. Two approaches can be applied to solving this unavailable gradient issue. 
One is employing a Straight Through Estimator (STE)~\cite{DBLP:journals/corr/BengioLC13,DBLP:journals/corr/abs-1903-05662} to set the gradient to the constant value of $1$ as
\begin{equation}
\begin{aligned}
&\textbf{Forward}: y=round(x) \\
&\textbf{Backward}: \frac{\partial y}{\partial x}=\textbf{1}_{x\in R}
\end{aligned}
,\end{equation}\label{eqn:STE}which is effective in the quantization training. 
The other approach employs Alternating Direction Method of Multipliers (ADMM) to iteratively solve the parameters with a target quantization scheme as the optimization contraint~\cite{leng2018extremely}, eliminating the need to backpropagate through the quantizer. 
In this work, we use a combination of ADMM and STE, as shown in Algorithm \ref{algo:ADMMQuantization}, which in general follows the ADMM algorithm for weight quantization and where the STE is only applied for activation quantization.}

\section{Sum-of-Power-of-2 (SP2) Quantization Scheme}\label{sec:SP2all}

\begin{table*}[ht]
\caption{\textbf{Analysis on the operations for weight-activation multiplication by two quantization schemes of the weights.}} \label{tab:operations}
\begin{center}
\begin{tabular}{c| c| c| c}
    \toprule
    &Weight & Activation & Ops for Weight $\times$ Activation\\
    \hline
    \hline
    \multirow{2}{*}{}Quantization Scheme & $m$-bit fixed-point & $n$-bit fixed-point & \multirow{2}{*}{$n$-bit addition for $m-2$ times}  \\
    \cline{1-3}
    Operands & $(m-1)$-bit integer & $n$-bit integer & ~ \\
    \hline
    \hline
    \multirow{3}{*}{}Quantization Scheme & $m$-bit SP2 & $n$-bit fixed-point & {shift by {up to $2^{m_1}-2$} bits}  \\
    \cline{1-3}
    \multirow{2}{*}{Operands} & $m_1$-bit integer, $m_2$-bit integer & \multirow{2}{*}{$n$-bit integer} & shift by {up to $2^{m_2}-2$} bits \\
    ~ & $m_1+m_2=m-1, m_1\geq m_2$ & ~ & up to $(n+2^{m_1}-2)$-bit addition \\
    \bottomrule
\end{tabular}
\end{center}
\end{table*}

{In this section, we propose
a new hardware-friendly {\em sum-of-power-of-2 (SP2) quantization scheme}, which enjoys the non-multiplication operations for the inference computation as the binary, ternary, and power-of-2 schemes, while achieving negligible 
inference accuracy degradation. }

\subsection{SP2 Quantization Scheme}

{The proposed hardware-friendly sum-of-power-of-2 (SP2) quantization scheme can be considered as a variant of the power-of-2 quantization. SP2 scheme can eliminate multiplication operations in the (quantized) DNN inference models (as the power-of-2 scheme), and at the same time
is designed to address the non-negligible accuracy loss of power-of-2 quantization.
This is achieved by solving the low precision issue in the tail ends of the weight distribution.}

\begin{figure}[t]
\centering  
\includegraphics[width=0.9\columnwidth]{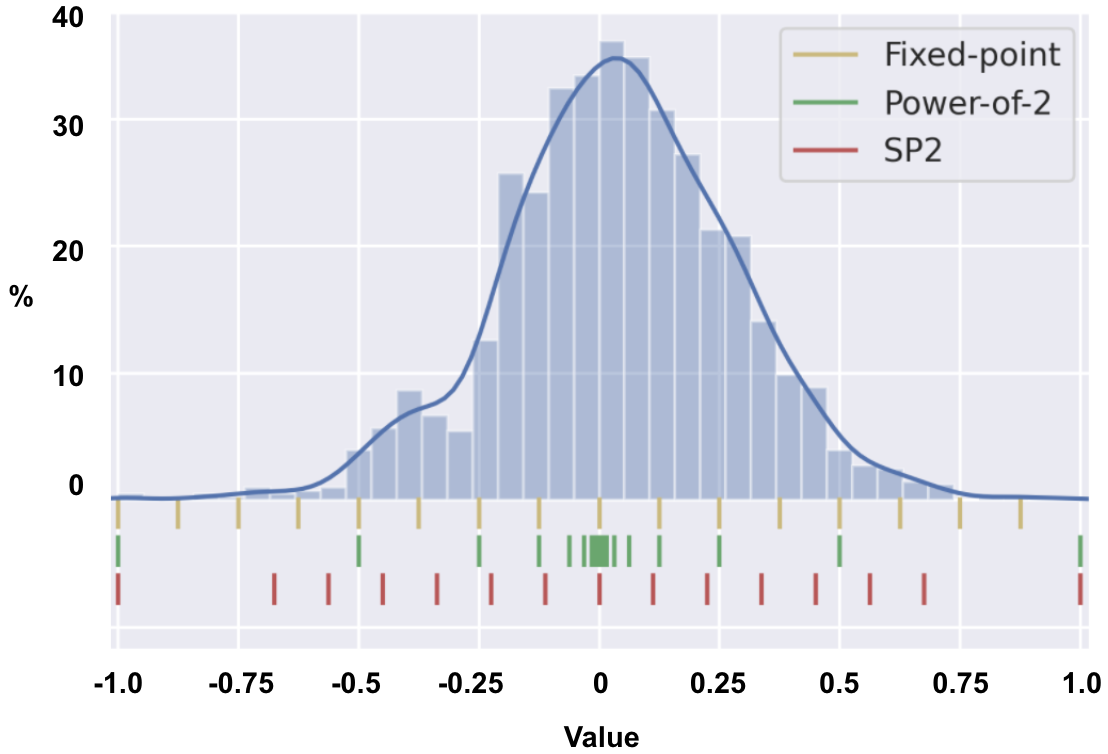}
\caption{\textbf{Quantization levels by fixed-point, power-of-2, and SP2 in 4-bit weight representation precision, and weight probability distribution of the 4th layer in MobileNet-V2.} 
}
\label{fig:distributions_sp2}
\end{figure}

{Formally, the quantized weight values by the sum-of-power-of-2 scheme with a total of $m$-bit representations are 
\begin{equation}\label{eq:SP2QL}
\begin{aligned}
& Q^{SP2}(m, \alpha)= \pm \alpha\times \{q_1+q_2\},\\
&  q_1 \in \{0, \frac{ 1}{2^{2^{m_1}-1}}, \frac{ 1}{2^{2^{m_1}-2}}, ...,  \frac{1}{2}\},
\\
& q_2 \in \{0, \frac{ 1}{2^{2^{m_2}-1}}, \frac{ 1}{2^{2^{m_2}-2}}, ...,  \frac{1}{2}\}
,
\end{aligned}
\end{equation}where $q_1$ and $q_2$ are power-of-2 numbers in similar format as the quantization levels in Eq. (\ref{eq:P2QL}), and $m_1$ and $m_2$ are the number of bits to represent the power-of-2 numbers i.e., $q_1$ and $q_2$, respectively.
Please note that with a total of $m$ bits to represent an SP2 quantized weight value, 1 bit is still reserved for the sign bit, and therefore we have $m_1+m_2+1=m$ with $m_1\geq m_2$.
In addition, the quantization levels by SP2 i.e., $\pm\{q_1+q_2\}$ are within $[-1,+1]$.
}

{Note that with $m$-bit representations, the SP2 scheme provides a total of $2^{m_1}\times2^{m_2}\times2-1=2^{m}-1$ quantization levels.
Although the power-of-2 quantization scheme in $m$-bit representations also provide $2^{m}-1$ quantization levels, the quantization levels resulted from the two schemes scatter distinctly, and therefore the schemes perform differently in preserving the accuracy of the quantized models.
In Figure \ref{fig:distributions_sp2}, the curve represents the actual probability distribution of DNN weights in a representative layer. 
Along the x-axis, we label the quantization levels by fixed-point, power-of-2, and our sum-of-power-of-2 quantization schemes. All the three schemes use $4$-bit representations and therefore each of them has 15 quantization levels within $[-1,+1]$.
}

{First, let us understand the intuition why
the power-of-2 quantized models incur the non-negligible accuracy degradation, while SP2 and fixed-point quantized models can achieve similar accuracy performance. 
The power-of-2 scheme has very high precision around the mean with only 4-bit weight presentation, but the tail ends present very low precision.
In contrast, our SP2 quantization possesses relatively evenly scattered quantization levels, which is close to that of fixed-point quantization levels, 
except the tail ends where very few weight values are presented. This explains the advantages of SP2 quantization scheme.}

\begin{table*}[h]
\caption{\textbf{Result from different quantization schemes for the ResNet-18 and MobileNet-v2 DNN models on CIFAR10, CIFAR100, and ImageNet datasets.}}\label{tab:comparison_schemes}
\centering
\begin{tabular}{c c c c c c}
\toprule
Quantization & Bit width & \multicolumn{2}{c}{ResNet-18 Accuracy (\%)} & \multicolumn{2}{c}{MobileNet-v2 Accuracy (\%)} \\
Scheme & (Wght./Actv.) & Top1 & Top5& Top1& Top5\\
\hline
\multicolumn{6}{c}{\textbf{CIFAR10}} \\
\hline
Baseline (FP) & 32/32 & 93.62 & - & 92.51 & -\\
P2 & 4/4 &92.97
($-$0.65)& - & 91.34($-$1.17)&-\\
Fixed & 4/4 & 93.43 ($-$0.19) & - & 92.34 ($-$0.17)& -\\
SP2 & 4/4 & 93.47
($-$0.15) & - & 92.72
($+0.21$)& -\\
\bf{MSQ} (half/half) & 4/4 & 93.53
($-$0.09)& - & 92.57
($+$0.06)& -\\
\bf{MSQ} (optimal) & 4/4 & 93.65
($+0.03$)&-  & 92.55
($+$0.04)&- \\
\hline
\multicolumn{6}{c}{\textbf{CIFAR100}} \\
\hline
Baseline (FP) & 32/32 &  74.49&  92.70& 71.48& 91.98\\
P2 & 4/4 &73.88
($-$0.61)&92.14
($-$0.56)&68.68
($-$2.80)& 90.06
($-$1.92)\\
Fixed & 4/4 &  74.37
($-$0.12)&  92.31 ($-$0.39)& 71.16
($-$0.32)& 91.63
($-$0.35)\\
SP2 & 4/4 &  74.33
($-$0.17)&  92.49
($-$0.21)& 71.13
($-$0.35)& 91.69
($-$0.29)\\
\bf{MSQ} (half/half) & 4/4 & 74.58
($+$0.09)& 92.39
($-$0.31) & 71.21
($-$0.27)& 91.74
($-$0.24)\\
\bf{MSQ} (optimal) & 4/4 & 74.60
($+$0.11)& 92.63
($-$0.07)& 71.50
($+$0.02)& 91.82
($-$0.16) \\
\hline
 & &\multicolumn{2}{c}{\textbf{ImageNet}} & \multicolumn{2}{c}{Wght./Actv. 4/32} \\
\hline
Baseline (FP) & 32/32 & 69.76 & 89.08 & 71.88& 90.29\\
P2 &4/4 &68.20
($-$1.56) &87.14
($-$1.94)& 69.93($-$1.95)& 88.63($-$1.66)\\
Fixed & 4/4 & 69.72
($-$0.04) &  88.67
($-$0.41)& 71.26
($-$0.62)& 90.18 ($-$0.11) \\
SP2 & 4/4 & 69.74
($-$0.02) &  88.71
($-$0.37)& 71.32
($-$0.56)& 90.17
($-$0.12)\\
\bf{MSQ} (half/half) & 4/4 &  70.11
($+$0.35)& 89.41
($+$0.33)& 71.26
($-$0.62)& 90.04
($-$0.25)\\
\bf{MSQ} (optimal) & 4/4 & {70.27} 
{($+$0.51)}& {89.42} 
{($+$0.34)}& 71.31($-$0.57) & 90.11($-$0.18) \\

\bottomrule
\end{tabular}
\end{table*}

{Next, we analyze the effect of SP2 quantization scheme on the computation of weight-activation multiplication. 
In Table \ref{tab:operations}, we compare fixed-point and SP2 quantization schemes of the weights, while throughout this paper we use fixed-point quantization for the activation.
In the first scheme with $m$-bit fixed-point quantization for the weight and $n$-bit fixed-point quantization on the activation, the weight operand is actually represented as the $(m-1)$-bit unsigned integer, since 1 bit is for the sign.
Although a quantization level is within $[-1,+1]$,  the actual weight operand is the $(m-1)$-bit unsigned integer.
And the activation operand is directly represented as the $n$-bit unsigned integer, because activations are non-negative.
The operations for implementing weight-activation multiplication are therefore $n$-bit additions for $(m-2)$ times.
}

{For the second scheme with $m$-bit SP2 quantization on the weight, we have an $m_1$-bit unsigned integer and an $m_2$-bit unsigned integer together to encode the quantization level of the quantized weight, and $m_1+m_2=m-1$ because 1 bit is for the sign. The quantization level is then $2^{b_1}+2^{b_2}$, where $b_1$ and $b_2$ are encoded with $m_1$ and $m_2$ bits, respectively. 
The weight-activation multiplication is implemented by (1) shift of the activation operand by $b_1$ bits, (2) shift of the activation operand by $b_2$ bits, and (3) addition of the two shifted operands.
Since $b_1$ and $b_2$ are encoded by $m_1$- and $m_2$-bit unsigned integer, respectively, Operations (1) and (2) can be shift by at most $2^{m_1}-2$ and $2^{m_2}-2$ bits, respectively. 
The shifted activation operands will be $n+2^{m_1}-2$ and $n+2^{m_2}-2$ bits respectively.
Therefore one $(n+2^{m_1}-2)$-bit addition is needed. 
In summary, with SP2 weight quantization, the weight-activation multiplication can be implemented with two shift operations and one addition operation.
}

\subsection{Accuracy Performance Analysis}\label{sec:P2FixedSP2accuracy}

{In this section, we discuss the accuracy performance of the fixed-point (Fixed), power-of-2 (P2), and proposed sum-of-power-of-2 (SP2) quantization schemes. Our baseline models use 32-bit floating-point (FP) for both weights and activations. All the quantization schemes apply 4-bit quantization. While different quantization schemes
are explored for weights, activations are using fixed-point quantization.
Table \ref{tab:comparison_schemes} summarizes the quantized models' accuracy with the accuracy changes (with respect to the baseline FP models) marked in the brackets. 
Experiments are conducted with ResNet-18 and MobileNet-v2 models on CIFAR10, CIFAR100, and ImageNet. 
As for activation quantization, 4-bit fixed-point is used for all the models, except the MobileNet-v2 on ImageNet dataset.
MobileNets are a family of specialized and lightweight models, therefore presenting unstable convergence under modifications such as pruning and quantization.
Activation quantization of MobileNet-v2 on ImageNet is performed in
\S\ref{sec:MSQcomparisonwithrefs} with comparisons to existing works.
}

In Table \ref{tab:comparison_schemes}, we now focus on P2, Fixed, and SP2 quantization schemes, and the MSQ scheme will be discussed in \S\ref{sec:MSQ}. 
First, power-of-2 (P2) results in significant accuracy degradation, around 1\%$\sim$2\% in general with extreme case of 2.80\% Top-5 accuracy loss of MobileNet-v2 on CIFAR100.
{For ImageNet, both the fixed-point (Fixed) and sum-of-power-of-2 (SP2) schemes have negligible accuracy loss, $\leq0.41$\% for ResNet-18 and $\leq0.62$\% for MobileNet-v2 accross the three datasets.
These two schemes achieve comparable accuracy of quantized models.
In summary, the 4-bit-width Fixed and SP2 quantization schemes are essentially equivalent in terms of the accuracy of the quantized models, and their accuracy losses are negligible.
}

\section{A FPGA-Centric Mixed Scheme Quantization }\label{sec:MSQ}

{In this section, we propose our mixed scheme quantization (MSQ) for FPGA implementations of DNN inference models. 
Based on the analysis in \S\ref{sec:P2FixedSP2accuracy}, fixed-point and SP2 quantization are equivalent in preserving the accuracy of the quantized models when with the same precision, e.g., 4-bit for both.
Therefore, in our proposed MSQ, the fixed-point and SP2 schemes with the 4-bit precision are applied in the DNN model quantization, (1) for better FPGA resource allocation, and (2) with negligible accuracy loss.
}

\subsection{Motivation}

{The idea of proposed mixed scheme quantization (MSQ) is to partition DNN weights in the same layer into two categories, one is handled by fixed-point quantization, and the other by SP2 quantization.
These two schemes use the same precision to facilitate hardware implementation. The motivations to adopt MSQ are: 
First, let a weight matrix be obtained by transforming the weight tensor of a layer into a 2D GEMM matrix with rows and columns. Weights in different rows of the matrix may present {\em different distributions}. 
For rows with more Gaussian-like weight distributions (with smaller variances), SP2 quantization is preferable; while 
for rows with more Uniform-like weight distributions (with larger variances), fixed-point quantization should be used.
Thus, the mixed scheme is necessary at {\em algorithm level}---it can achieve similar or even potentially
higher accuracy than existing schemes.
Second, our approach also leads to a better utilization of 
heterogeneous resources available in FPGA---
weights based on the two schemes can be managed by 
LUT and DSP resources. 
Specifically, the operations involving SP2 quantized weights should be implemented by LUTs; while those with fixed-point quantized weights can leverage the DSPs, the more limited resources on FPGA for DNN hardware accelerators. 
Overall, our MSQ achieves a sweet design spot achieving both high accuracy and 
processing throughput, thanks to the high and optimized
utilization of both LUTs and DSPs.  }

\begin{algorithm}[t]
\small
\caption{\small FPGA-Centric Mixed Scheme Quantization(MSQ) }\label{algo:MSQQuantization}
\SetKwInOut{Input}{input}
\SetKwInOut{Output}{target}
\SetKwFunction{SGD}{SGD}
\Input{
\noindent 32-bit floating-point DNN model $\mathcal{M}$, with weights  $\mathbf{W}$ to be quantized.\\

}
\Output{ Quantized model $\hat{\mathcal{M}}$ }
\BlankLine
\tcp{Initialization:} 
$U^0 = 0$;
$Z^0 = \mathbf{W}$;\\
Partition rate  $PR_{SP2}$ from FPGA resource characterization;\\
$\mathbb{S}_f = $Fixed-point;
$\mathbb{S}_p = $SP2;
\BlankLine

\ForEach{Epoch}{

Calculate variance $v_r^{(l)}$ for each $r$-th row of the layer $l$ weight matrix $\mathbf{W}^{(l)}$;\\
Sort $v_{1:R}^{(l)}$ to obtain the threshold $\theta^{(l)}$ such that $PR_{SP2}$ of the rows with variances less than $\theta^{(l)}$;\\

   
\textbf{if} $v_r^{(l)} < \theta^{(l)}$ \textbf{then} $\mathbb{S} \leftarrow \mathbb{S}_p$\;\textbf{else} $\mathbb{S} \leftarrow \mathbb{S}_f$\;

\tcp{Update $Z$, $U$:}

$Z^{t}\leftarrow$ $\textbf{proj}_\mathbb{S}$($\mathbf{W}+U^{t-1}$);\\

$U^{t}\leftarrow \mathbf{W}-Z^{t}+U^{t-1}$;\\

\ForEach{Batch}{

$input \leftarrow \textbf{proj}_\mathbb{S}(input)$;\\ 
$loss\leftarrow \mathcal{M}(input)$;\\
$loss\leftarrow loss+\sum \frac{1}{2} \rVert \mathbf{W}-Z^t+U^t \rVert^2$;\\
Backpropagate $loss$ and update $\mathbf{W}$;\\
\BlankLine
}
}
Return $\hat{\mathcal{M}} \leftarrow \mathcal{M}\{\textbf{proj}_\mathbb{S}(\mathbf{W})\}$.
\end{algorithm}

\subsection{Algorithm}\label{sec:MSQ_alg}

{In MSQ, 
each row in a weight matrix should employ either the SP2 or  fixed-point scheme.
To determine the  scheme for each row, the weight variances of all the rows are calculated.
{We define a threshold $\theta$ for the variances, such that for the rows with smaller variances than the threshold, the SP2 quantization is employed; and otherwise, the fixed-point scheme is applied. By setting the proper threshold $\theta$, the desired partition ratio of SP2 to fixed-point can be achieved with improved FPGA resource utilization.}
Algorithm \ref{algo:MSQQuantization} provides the details.

The optimal ratio of SP2 to fixed-point is determined by the available resources on FPGA devices and resource utilization required to support the design. Generally, the utilization factor of DSPs should be maintained at 100\% to take full advantage of the DSP resource for the fixed-point multiplications. 
When only fixed-point quantization is applied, the LUT utilization is low even though DSP utilization reaches the maximum. Incorporating the SP2 quantization can increase the LUT utilization, and therefore enhancing the throughput. The exploration of the optimal ratio of SP2 to fixed-point 
among the weight matrix rows is elaborated in §\ref{sec:FPGAevaluation}. 
}

\subsection{Accuracy Results}\label{sec:MSQcomparisonwithrefs}

\subsubsection{Experiment Setup}

{We evaluate our  MSQ in three application domains i.e., image classification with convolutional neural networks (CNNs); object detection and recognition with YOLO-v3; machine translation, speech recognition, and sentiment classification with recurrent neural networks (RNNs). 
We use no extra data augmentations in our quantization, other than those already employed for training the 32-bit floating-point baseline models. 
Our quantization training algorithm uses step or cosine learning rate decay and $\ell_2$ regularization, following training algorithms of the baseline models. Our quantization algorithms are implemented with the PyTorch framework on NVIDIA TITAN RTX GPUs and GeForce RTX 2080Ti GPUs.}

{
For image classification, we evaluate with the deep residual net (ResNet-18)~\cite{he2016deep}, which is a widely used model for computer vision tasks, as well as the lightweight MobileNet-v2 model \cite{sandler2018mobilenetv2}.
We test on CIFAR10 \cite{krizhevsky2009learning}, CIFAR100 \cite{krizhevsky2009learning}, and ImageNet ILSVRC-2012 \cite{krizhevsky2012imagenet} datasets. DNN models for CIFAR10 and CIFAR100 datasets are trained from scratch and quantized for $150$ epochs.
For ImageNet dataset, pre-trained models  in 32-bit floating-point are used and quantized for $90$ epochs. 
The initial learning rate is $8e-3$ for CIFAR10, $4e-3$ for CIFAR100, $5e-4$ for ImageNet.}

{
For object detection, we explore the implementation of a fully convolutional neural network (FCNN) called YOLO-v3 \cite{DBLP:journals/corr/abs-1804-02767} on MS COCO 2014 \cite{DBLP:journals/corr/LinMBHPRDZ14} dataset. 
The learning rate starts from $1e-2$, and decays to $5e-4$ with cosine annealing. We evaluate mean Average Precision (mAP) at an IoU threshold value of 0.5 (mAP$@0.5$), as well as average mAP over the IoU threshold range from 0.5 to 0.95 (mAP$@(0.5:0.95)$). }

{For RNNs, we evaluate three networks. The first one is an LSTM network with $256$ hidden neurons in two layers~\cite{hochreiter1997long} on Penn Tree Bank (PTB)~\cite{marcus1993building} dataset for the machine translation application with perplexity (PPL) as the evaluation metric (lower PPL is better). 
The second is a network based on GRU with $1024$ hidden neurons in two layers~\cite{cho2014learning} on TIMIT acoustic-phonetic continuous speech corpus~\cite{garofolo1993darpa} dataset for the speech recognition application. The evaluation metric is Phoneme Error Rate (PER) and lower PER is better. 
Finally, we use another LSTM network with three hidden layers each having $512$ neurons on IMDB \cite{maas2011learning} dataset for sentiment classification. 
Our learning rate is $1e-3$ for all the RNNs. }

\begin{table}[t]
\caption{\textbf{Comparisons with existing works with ResNet-18 model on ImageNet dataset.}}
\centering
\begin{tabular}{cccc}
\toprule
\multirow{2}{*}{Methods} & Bit-width & \multirow{2}{*}{Top-1 (\%)} & \multirow{2}{*}{Top-5 (\%)}\\
~ & (W/A) & ~ & ~ \\
\hline
Baseline(FP) & 32/32 & 69.76 & 89.08\\
Dorefa\cite{zhou2016dorefa} & 4/4 & 68.10 & 88.10\\
PACT \cite{choi2018pact} & 4/4 & 69.20 & 89.00\\
DSQ\cite{gong2019differentiable} & 4/4 & 69.56 & N/A\\
QIL\cite{jung2019learning} & 4/4 & 70.10 & N/A\\
$\mu$L2Q\cite{cheng2019uL2Q} & 4/32 & 65.92 & 86.72\\
LQ-NETS\cite{zhang2018lq} & 4/4 & 69.30 & 88.80\\
\textbf{MSQ} &\textbf{4/4} & \textbf{70.27} & \textbf{89.42}\\

\bottomrule
\end{tabular}
\label{tab:imagenetresnet}
\end{table}



\begin{table}[t]
\caption{\textbf{Comparisons with existing works with MobileNet-v2 model on ImageNet dataset.}}
\centering
\begin{tabular}{c c c c}
\toprule
\multirow{2}{*}{Methods} & Bit-width & \multirow{2}{*}{Top-1 (\%)} & \multirow{2}{*}{Top-5 (\%)} \\
~ & (W/A) & ~ & ~ \\
\hline
Baseline(FP) & 32/32 & 71.88 & 90.29\\
PACT\cite{choi2018pact} & 4/4 & 61.40 & N/A\\
DSQ\cite{gong2019differentiable} & 4/4 & 64.80 & N/A\\
{\textbf{MSQ}} & \textbf{4/4} &  \textbf{65.64}&  \textbf{86.98}\\
\bottomrule
\end{tabular}
\label{tab:imagenetmbnet}
\end{table}

\begin{table}[t]
\caption{\textbf{YOLO-v3 on COCO 2014 dataset with 4-bit  quantization.} ($8\times$ compression rate)}

\centering
\begin{tabular}{c| c| c| c}
    \toprule
    Image Size& Scheme & mAP $@0.5:0.95$ &mAP $@0.5$\\
    \toprule
    \multirow{2}{*}{320}&Baseline(FP)&37.7&56.8\\
    &MSQ&35.8&53.9\\
    \hline
    \multirow{2}{*}{640}&Baseline(FP)&45.6&64.7\\
    &MSQ&44.1&64.8\\
    \bottomrule
\end{tabular}
\label{tab:yolo}
\end{table}

\begin{table}[t]
\caption{\textbf{RNN on machine translation, speech recognition, and sentiment classification.}} 
\centering
\begin{tabular}{c c c c}
\toprule
\multirow{2}{*}{Scheme} & Bit Width &\multirow{2}{*}{Evaluation Metric}& \multirow{2}{*}{Result}\\
\multirow{2}{*}{} & (W/A)& \multirow{2}{*}{}&\multirow{2}{*}{} \\
\toprule
\multicolumn{4}{c}{\textbf{LSTM on PTB}}\\
\hline
EQMBaseline(FP) & 32/32 & Perplexity& 109\\
 &  & (PPL) lower  better& \\
EQM\cite{he2016effective} & 4/4 & PPL& 114\\
\noalign{\global\arrayrulewidth=0.1pt}
\hline
\noalign{\global\arrayrulewidth=0.4pt}
OurBaseline(FP) & 32/32 & PPL& 110.89\\
Fixed & 4/4 & PPL& 113.03\\
SP2 & 4/4 & PPL&113.42\\
\textbf{MSQ(half/half)} & 4/4 &  PPL&112.74\\
\textbf{MSQ(optimal)} & 4/4 &  PPL&112.72\\
\toprule
\multicolumn{4}{c}{\textbf{GRU on TIMIT}}\\
\hline
OurBaseline(FP) & 32/32 & Phoneme Error Rate& 19.24\%\\
&&(PER) lower  better&\\
Fixed & 4/4 & PER& 20.14\%\\
SP2 & 4/4 & PER& 20.09\%\\
\textbf{MSQ(half/half)} & 4/4 &  PER& 19.58\%\\
\textbf{MSQ(optimal)} & 4/4 &  PER&19.53\% \\
\toprule
\multicolumn{4}{c}{\textbf{LSTM on IMDB}}\\
\hline
EQMBaseline(FP) & 32/32 & Accuracy& 89.54\%\\

EQM\cite{he2016effective} & 4/4 & Accuracy& 88.47\%\\
\noalign{\global\arrayrulewidth=0.1pt}
\hline
\noalign{\global\arrayrulewidth=0.4pt}
OurBaseline(FP) & 32/32  & Accuracy& 86.37\%\\
Fixed & 4/4 & Accuracy& 86.12\%\\
SP2 & 4/4 & Accuracy& 86.02\%\\
\textbf{MSQ(half/half)} & 4/4 &  Accuracy& 86.28\%\\
\textbf{MSQ(optimal)} & 4/4 &  Accuracy& 86.31\% \\
\toprule
\end{tabular}
\label{tab:rnn}
\end{table}

\subsubsection{Result Analysis}

{Tables \ref{tab:comparison_schemes}, \ref{tab:imagenetresnet}, and \ref{tab:imagenetmbnet} summarize quantization results for the image classification.
Table \ref{tab:comparison_schemes} compares different quantization schemes including power-of-2 (P2), fixed-point (Fixed), sum-of-power-of-2 (SP2), and our mixed scheme quantization (MSQ). 
Two partitioning ratios are tested for MSQ, the first one being $PR_{SP2:Fixed}=1:1$, and the second one being $PR_{SP2:Fixed}=2:1$ that is the optimal ratio  from FPGA characterizations. 
On Top-1 accuracy, MSQ has the minimum accuracy loss for most cases. 
}

{
The accuracy increase of MSQ compared to sole SP2 or Fixed results from several aspects. First, combining SP2 and Fixed makes the quantized DNN weights fit the original weight distribution better.
In addition, model compression could slightly increase accuracy when weight bit-width $\geq 4$, as the quantization resolution is high enough so that the inference results of DNNs are not affected, and quantization noise can potentially act as regularization that benefits generalization and addresses overfitting.}

{Tables \ref{tab:imagenetresnet}, 
and \ref{tab:imagenetmbnet} compare our MSQ with existing DNN quantization works including Dorefa \cite{zhou2016dorefa},
PACT \cite{choi2018pact},
DSQ \cite{gong2019differentiable},
QIL \cite{jung2019learning},
$\mu$L2Q \cite{cheng2019uL2Q}, and
LQ-NETS \cite{zhang2018lq}.
Those works and our MSQ start with the same pre-trained models with the same baseline accuracy.
Here we use the optimal $PR_{SP2:Fixed}=2:1$ in MSQ. 
Note that this optimal ratio is from hardware characterization, not for increasing accuracy.
Table \ref{tab:imagenetresnet} shows that Dorefa, PACT, DSQ, $\mu$L$2$Q, and LQ-NETS have up to $3.84\%$ accuracy degradation, only QIL reports lossless accuracy performance. Our MSQ increases accuracy by $0.49\%$  compared with the floating-point model.}
Table \ref{tab:imagenetmbnet} shows that the lightweight model MobileNet-v2 is much harder to quantize with 4-bits (for both weight and activation), our MSQ achieves the highest accuracy of the quantized models. 

{On the even larger YOLO-v3 model for object detection, we apply 4-bit quantization, which is equivalent to 8$\times$ compression rate. We test on two image sizes i.e., 320$\times$320 and 640$\times$640.
Our MSQ performs very well in preserving the mAP values i.e., with negligible mAP degradation, and for the case of 640$\times$640 input size and mAP$@0.5$, MSQ can even increase the mAP value. We notice a slightly higher mAP degradation  when the input size is small. 
This is because the 
smaller feature maps are more sensitive to quantization error. 
There is no existing quantization methods reporting about YOLO network quantization.
To provide an idea about the mAP degradation by our MSQ, we can compare with the weight pruning method on YOLO with also 8$\times$ compression rate \cite{DBLP:journals/corr/abs-1907-11093}, which decreases the mAP$@0.5$ by $\sim3.0$ on a simpler dataset than COCO 2014.
In general, at the same compression rate, and especially when the dataset is simpler, weight pruning should have less accuracy degradation than weight quantization. 
But our MSQ can have comparable or even smaller mAP degradation.
It demonstrates our MSQ works very well on YOLO networks.}

{Table \ref{tab:rnn} shows that our MSQ scheme outperforms the Fixed and SP2 quantization for all the three RNN tasks. 
We also compare our method with existing work EQM\cite{he2016effective} on the PTB and IMDB datasets.
Because we do not have the same pre-trained models as in EQM \cite{he2016effective}, we also need to report on their pre-trained (32-bit floating-point) baseline models.
On the PTB dataset, EQM \cite{he2016effective} increases perplexity (PPL) by 5.00 (the lower the better), while our MSQ only increases by $<2.00$. For the IMDB dataset, EQM loses near $1\%$ accuracy and MSQ only loses $0.06\%$ accuracy.}
Note that we have not found any DNN quantization works investigating the TIMIT dataset, so we could not compare with existing works on TIMIT. 
\section{FPGA Implementation: Design and Optimization}\label{sec:FPGA}

{Besides obtaining accuracy advantage, the proposed MSQ assembling the fixed-point and SP2 quantization schemes significantly promotes the efficiency of the FPGA deployment. Specifically, the newly joined SP2 quantization provides} two apparent advantages in the hardware aspect:
(i) the multiplication arithmetic {involving the SP2 quantized weights can be implemented with} 
simple logic shifter and adder, instead of the conventional multiplier; and
(ii) since the FPGA underlying components include DSP and LUT, the rest LUTs can be leveraged for
{computations with SP2 weights} 
while the DSPs {are simulatenously} fully utilized for conventional multiplication. 
{Therefore, with the proposed MSQ as an {\em ensemble} of fixed-point and SP2,}
the same device can possibly deliver higher performance than existing designs, in which the throughput is theoretically bounded by the DSP count. 

{This section} addresses the hardware design challenges with mixed number systems. 
Please note that the hardware benefit from SP2 is orthogonal to prior research efforts (e.g., dataflow~\cite{sun2019power} and locality~\cite{guan2017fp} optimization), and therefore can be employed by {\em any} existing DNN accelerator.

\subsection{{FPGA Resource Characterization}}

\begin{figure}[htb]
  \centering
  \includegraphics[width=0.85\columnwidth]{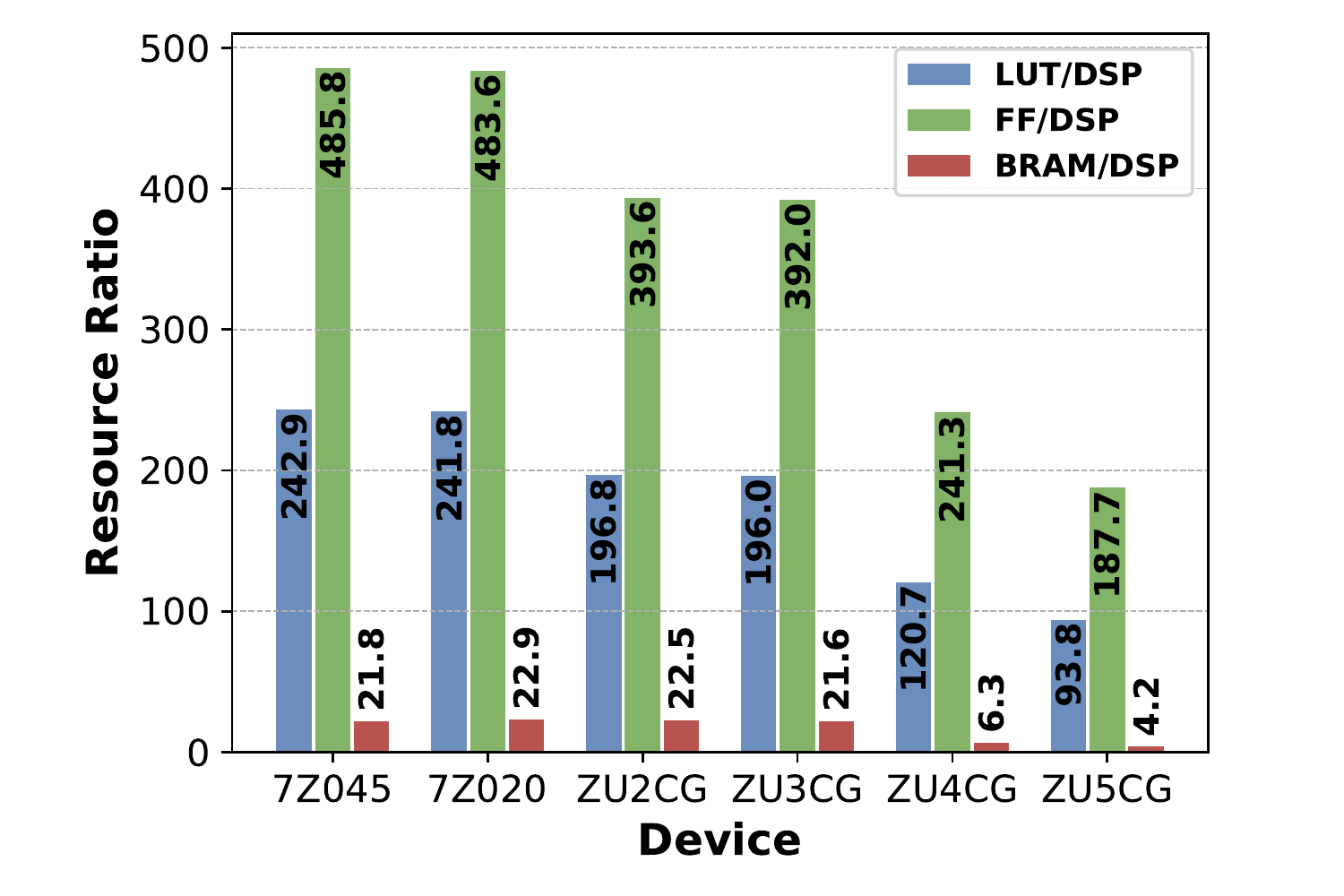}
  \caption{\textbf{Resource ratio of different FPGA devices.} For each device, LUT, FF, and BRAM numbers are all normalized with respect to DSP number.}
  \label{fig:deviceresource}
\end{figure}

\begin{figure*}[t]
  \centering
  \includegraphics[width=0.8\textwidth]{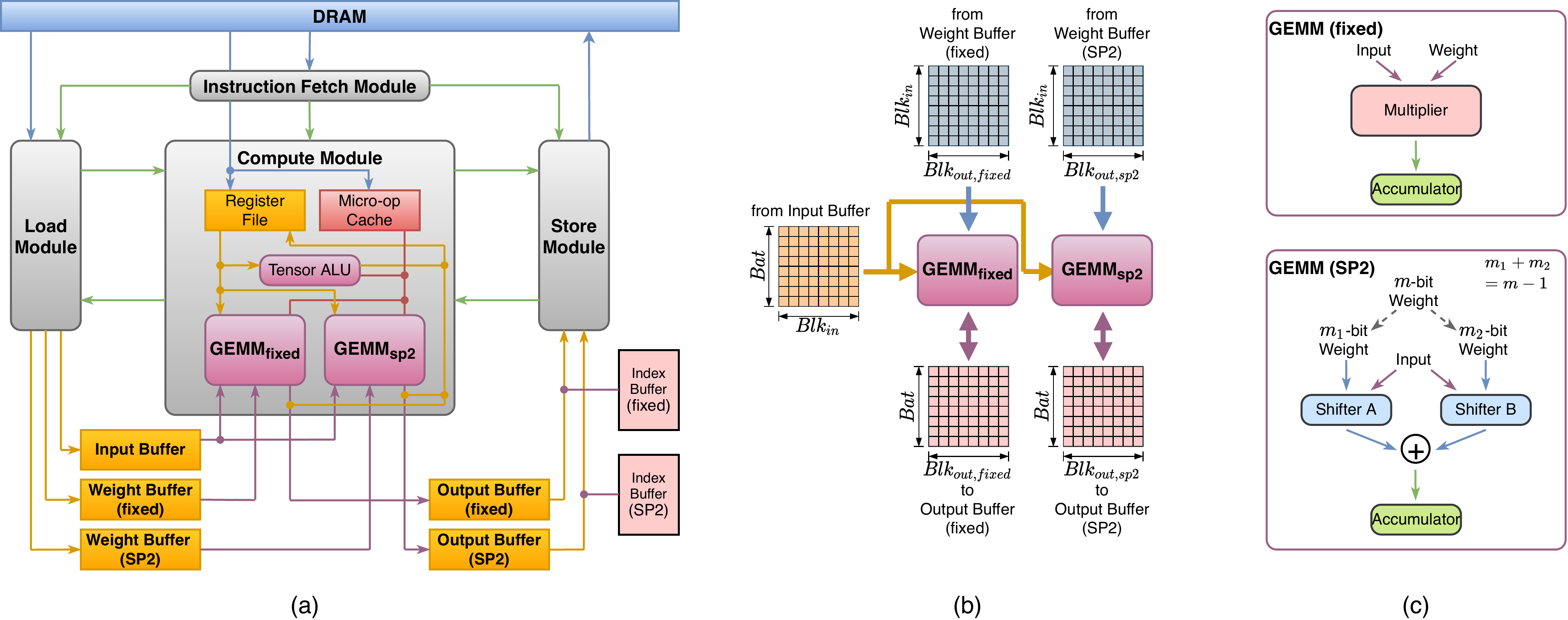}
  \caption{\textbf{Hardware architecture of convolution for MSQ number system.} (a) Overall framework with $\hw{GEMM}_{fixed}$ core for fixed-point operations and $\hw{GEMM}_{fixed}$ core for SP2 operations; (b) Dataflow in heterogeneous \hw{GEMM} cores; and (c) Computations in heterogeneous \hw{GEMM} cores. }
  \label{fig:arch}
\end{figure*}

FPGA devices provide different types of resource{s}, i.e., DSP, LUT, BRAM, and FF, for computation and storage, and the resource amount ratios vary in different {FPGA} devices. 
Figure~\ref{fig:deviceresource} presents the resource ratios of Zynq series devices (each device name starts with ``XC'' that is omitted for simplicity), with each bar normalized by the DSP count on the corresponding device.
The ratio of LUT to DSP attracts our attention, since this number directly decides the building block for multiplications with fixed-point and SP2 quantized weights, respectively. 
Apparently, the ratio of LUT/DSP in XC7Z045/XC7Z020 devices are larger than that in XCZU4CG/XCZU5CG devices. This also occurs in FPGA devices of other types.
{Specifically, since the multiplications with fixed-point and SP2 weights consume the DSP and LUT, respectively, the LUT/DSP ratio decides the parallel PE counts for these two operation types. 
For different devices, we select different proper ratios of PE counts for fixed-point and SP2 according to the available resource amount. 
Importantly, the PE ratio is used as the desired SP2/fixed-point ratio and sent to Algorithm~\ref{algo:MSQQuantization} to obtain the properly quantized models with the novel MSQ scheme.}

\subsection{Architecture with Heterogeneous GEMM Engines}

This section provides a design based on the versatile tensor accelerator (VTA)~\cite{moreau2019hardware}. 
The hardware framework contains four modules as shown in Figure~\ref{fig:arch}(a), where the \hw{Instruction} module loads the instructions and provides control signals to other modules.
\hw{Load} and \hw{Store} modules control the input/output activation and weight data communication between on-chip buffers and DRAM.
The \hw{Compute} module executes the workloads, with the \hw{RegFile} as the scratchpad memory for partial sum accumulation and \hw{TensorALU} computing the element-wise operations (e.g., activation).
The major computation components are the general purpose matrix multiplication (GEMM) cores. Different from VTA, there are two heterogeneous \hw{GEMM} cores, \hw{GEMM_{fixed}} for conventional multiplications, and \hw{GEMM_{sp2}} for SP2 operations. 
Besides conventional GEMM acceleration framework, our \hw{GEMM_{fixed}} can be naturally combined with advanced GEMM acceleration frameworks with architectural optimizations on the fixed-point operations (and uses DSP resources on FPGA). An example is Bit-Fusion \cite{sharma2018bit}, which is orthogonal and can be combined with our MSQ. Firstly, the fixed point operations executed on DSP in our MSQ framework can be accelerated by Bit-Fusion. Secondly, MSQ assigns a large portion (beyond 50\%) of computations in each layer to SP2 and leverages LUTs for computation, which are previously not fully exploited by fixed-point acceleration techniques like Bit-Fusion. 
A doubling performance can be anticipated as fixed-point and SP2 are computed in parallel on FPGA.

The detailed workflow of two \hw{GEMM} cores is illustrated in Figure~\ref{fig:arch}(b). 
A tiled block of input activation data with a size of $Bat\times Blk_{in}$ is read from the input buffer to the register array, where $Bat$ is the batch size and $Blk_{in}$ is the input channel count of the tile that will be computed in parallel. 
Note that the input activation will be broadcasted to both \hw{GEMM} cores. 
As Figure~\ref{fig:arch}(c) displays, the \hw{GEMM_{fixed}} core is composed of multipliers implemented with DSPs on FPGA, while the \hw{GEMM_{sp2}} uses LUTs to realize shift and addition for the novel SP2 based computations. 
Meanwhile, two weight buffers provide the weight values in fixed-point and SP2 formats, respectively. 
The partial results will be accumulated and stored in individual register filers, and the final results are written to individual output buffers. 
Because the filters are allocated to heterogeneous \hw{GEMM} cores depending on their weight representation format, two filter index buffers are set to instruct the \hw{Store} unit to write the output data to the proper global addresses.
Figure~\ref{fig:arch}(c) gives a detailed structure to handle fixed-point and SP2 operations in two \hw{GEMM} cores.

Two design parameters $Blk_{out,fixed}$ and $Blk_{out,sp2}$ indicate the parallel PE count in each \hw{GEMM} core and size of corresponding registers array, as illustrated in Figure~\ref{fig:arch}(b).
Two factors are considered in selecting $Blk_{out,fixed}$ and $Blk_{out,sp2}$.
One is that the ratio of $Blk_{out,fixed}$ to $Blk_{out,sp2}$ should be equivalent to that of different weight types (fixed-point/SP2).
An imbalanced ratio may result in under-utilization of the certain \hw{GEMM} core. 
The other is that the on-chip resources (DSP and LUT count) should yield a particular ratio of design parameters, i.e., a proper number facilitates fully utilization of FPGA resources, which is the key motivation of this work. 
Specifically, we develop an FPGA-centric MSQ quantization method as mentioned in \S\ref{sec:MSQ_alg} that automatically trains quantized DNN models to achieve a particular ratio that meets the resource allocation on FPGA devices.
Additionally, we incorporate the processing operations after the convolution computations into the GEMM cores, including batch normalization, activation (ReLU) and pooling, as these operations consume few resources and incur negligible latency increase compared with convolution computations.
\section{Evaluation}\label{sec:FPGAevaluation}

\subsection{Experiment Setup}

To demonstrate the improvement of the proposed SP2 (and MSQ) scheme in the hardware aspect, we implemented the architecture with heterogeneous \hw{GEMM} cores on the embedded FPGA device, in which a high efficiency is usually in demand under resource limitation. 
As Table~\ref{tab:impl_param} shows, we implemented the architecture on the Zynq XC7Z020 and XC7Z045 devices with different design parameters that result in different throughput and resource utilization results.
Note that for each device, we set up different ratios between the PE array sizes of the \hw{GEMM_{fixed}} and \hw{GEMM_{sp2}} cores. 
Specifically, we progressively increase the size of \hw{GEMM_{sp2}} core ($Blk_{out,sp2}$), till the LUT utilization reaches $70\%$. 
For example, on XC7Z020 the desired fixed/SP2 ratio is 1:1.5 and on XC7Z045 it is 1:2.
For all implementations, the quantization bit-width for the CNN models is fixed to 4, and the working frequency is set to 100MHz.

\begin{table}[t]
\caption{\textbf{Hardware implementation parameters with different devices and settings. }
$Bat$, $Blk_{in}$, and $Blk_{out,fixed}$ are set such that the DSP utilization could reach maximum. $Blk_{out,sp2}$ is increased until the LUT utilization is high enough and optimized.}
\label{tab:impl_param}
\setlength\tabcolsep{2.5pt}
\centering
\begin{tabular}{c|c|cccc|c|c}
    \toprule
    \multirow{2}{*}{Impl.} & \multirow{2}{*}{Device} & \multirow{2}{*}{$Bat$} & \multirow{2}{*}{$Blk_{in}$} & \multicolumn{2}{c|}{$Blk_{out}$} & Ratio & Peak Thrpt. \\
    ~ & ~ & ~ & ~ & Fixed & SP2 & (fixed/SP2) & (GOPS) \\
    \hline
    D1-1 & \multirow{3}{*}{XC7Z020} & 1 & 16 & 16 & 0 & 1:0 & 52.8 \\
    D1-2 & ~ & 1 & 16 & 16 & 16 & 1:1 & 106 \\
    D1-3 & ~ & 1 & 16 & 16 & 24 & 1:1.5 & 132 \\
    \hline
    D2-1 & \multirow{3}{*}{XC7Z045} & 4 & 16 & 16 & 0 & 1:0 & 208 \\
    D2-2 & ~ & 4 & 16 & 16 & 16 & 1:1 & 416 \\
    D2-3 & ~ & 4 & 16 & 16 & 32 & 1:2 & 624 \\
    \bottomrule
\end{tabular}
\end{table}

\subsection{Evaluation with FPGA}

\subsubsection{Resource Utilization}
\label{sec:utilization}

\begin{figure}[htp]
  \centering
  \includegraphics[width=0.40\textwidth]{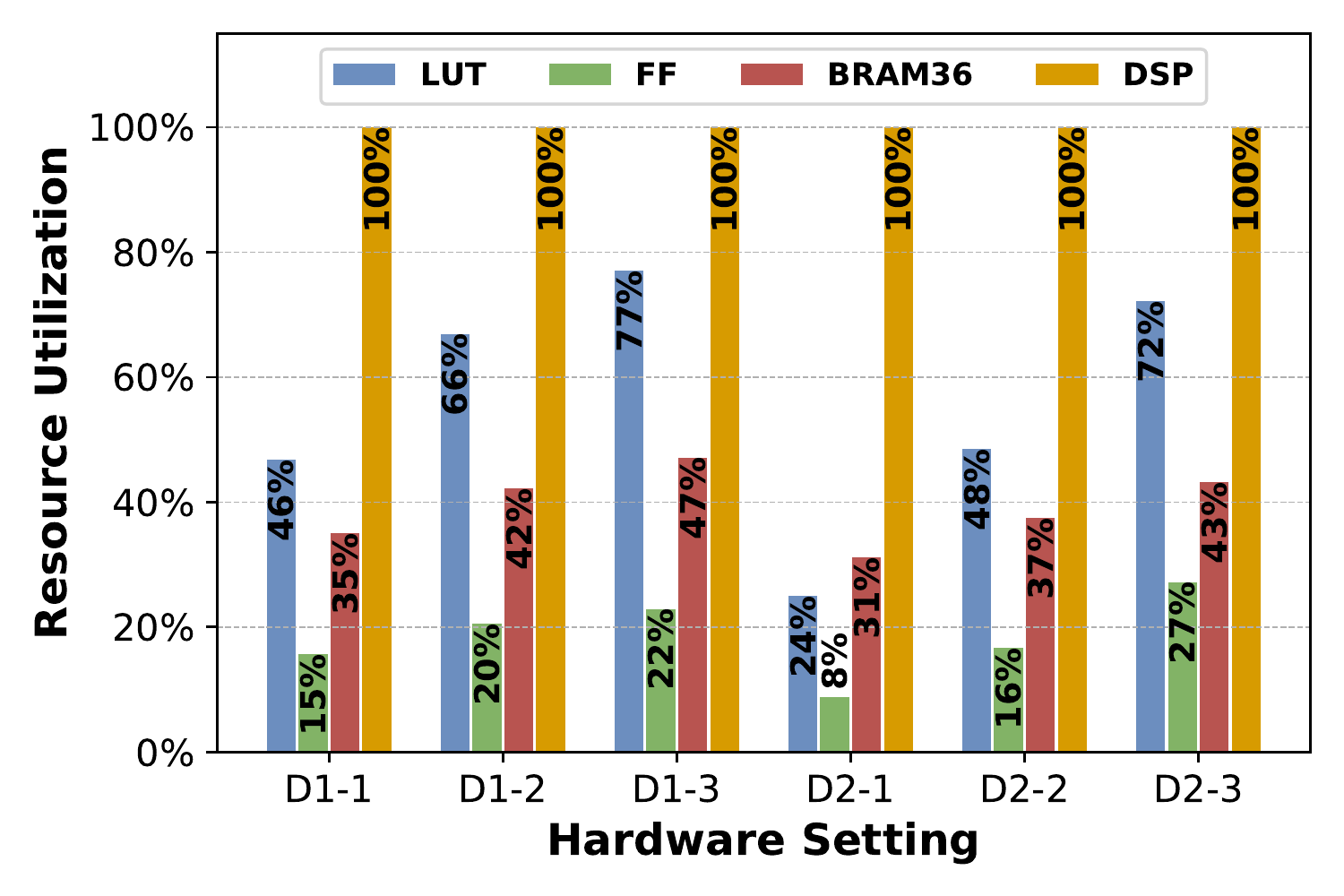}
  \caption{\textbf{FPGA resource utilization with different devices and settings. }In the three designs for each of the two devices, the DSP utilization is maintained at 100\% and the LUT utilization is raised to 70\%$-$80\% with FF and BRAM resources.}
  \label{fig:utilization}  
\end{figure}

Figure~\ref{fig:utilization} presents the resource utilization with different implementations.
Apparently, with the size increase of \hw{GEMM_{sp2}}, more on-chip LUT can be leveraged for a better peak throughput (GOPS).
As Table~\ref{tab:impl_param} shows, on XC7Z020 device (D1-1,2,3), the peak throughput was improved to $2.5\times$ (from 52.8 to 132 GOPS) with the extra \hw{GEMM_{sp2}} core. 
The maximum size of the PE array for SP2 is $1.5\times$ of that for fixed point. 
This peak throughput improvement is $3\times$ on XC7Z045, from 208 to 624 GOPS.
Although the ratio of available LUT/DSP is the same between the two devices, the optimal proportion of PE count for SP2 on XC7Z020 ($1.5\times$ of fixed-point) is a smaller than that on XC7Z045 ($2\times$ of fixed-point). 
This is because a portion of LUTs is consumed by \hw{Load} and \hw{Store} modules to accommodate the \hw{GEMM_{sp2}} core. 
The proposed architecture design is general for all devices through adjusting the $Blk_{out,sp2}$ to fully utilize the LUT resource and quantizing the models with the corresponding fixed-point/SP2 ratio. 

\subsubsection{Real-world Performance and Comparison}

To present the performance with real-world applications, we employed different CNN and RNN models with the proper SP2/fixed-point ratios on the two devices. 
The networks ResNet-18 and MobileNet-v2 are implemented based on the ImageNet dataset.
The performance results of each network under various hardware configurations are displayed in Table~\ref{tab:perf}. 
For some layers in CNNs like the first convolutional layer, the peak throughput cannot be reached since the number of input channels is less than $Blk_{in}$ so that the data cannot fill all of the PEs.
Generally, for CNN models, the overall PE utilization reaches $52.4\%$ to $70.1\%$, and the heterogeneous \hw{GEMM_{fixed}} and \hw{GEMM_{sp2}} cores improve the throughput by $2.1\times-2.5\times$ with the optimal design compared to utilizing the \hw{GEMM_{fixed}} core only. Compared with the design with only 4-bit fixed-point (fixed4/SP2 $=1:0$) quantization, the optimal design with the ratio of fixed4/SP2 $=1:1.5$ on XC7Z020 decreases the latency per image from 100.7ms to 47.1ms (2.13$\times$) for ResNet-18, and the optimal design with the ratio of fixed4/SP2 $=1:2$ on XC7Z045 decreases the latency from 25.1ms to 10.1ms (2.49$\times$) for ResNet-18. The latency improvement is more significant when compared with the 8-bit fixed-point design, as the optimal design on XC7Z020 achieves latency decrease from 181.3ms to 47.1ms (3.83$\times$), and the optimal design on XC7Z045 achieves latency decrease from 45.2ms to 10.1ms (4.48$\times$). As for RNN models, the PE utilization is $42.9\%-59.2\%$, and the performance is increased by $2.4\times-4.1\times$.

\begin{table*}[htb]
\caption{\textbf{Performance of various DNN applications on hardware under different settings.}}
\label{tab:perf}

\centering
\begin{tabular}{c|c|cccc|cc|c|ccc}
    \toprule
    \multirow{3}{*}{Device} & \multirow{3}{*}{\makecell{Ratio \\ (fixed/SP2)}} & \multicolumn{4}{c|}{Utilization} & \multicolumn{6}{c}{Throughput (GOPS)} \\
    \cline{3-12}
    ~ & ~ & \multirow{2}{*}{LUT} & \multirow{2}{*}{DSP} & \multirow{2}{*}{BRAM36} & \multirow{2}{*}{FF} & ResNet-18 & MobileNet-v2 & YOLO-v3 & LSTM & GRU & LSTM \\
    ~ & ~ & ~ & ~ & ~ & ~ & on ImageNet & on ImageNet & on COCO & on PTB & on TIMIT & on IMDB \\
    \hline
    \multirow{3}{*}{XC7Z020} & 1:0 & 12160 & 220 & 39 & 9403 & 36.0 & 33.0 & 36.6 & 26.1 & 22.6 & 25.0 \\
    ~ & 1:1 & 22912 & 220 & 49 & 14523 & 74.4 & 65.7 & 74.1 & 52.9 & 49.2 & 58.7 \\
    ~ & 1:1.5 (opt.) & 28288 & 220 & 56 & 17083 & 77.0 & 71.8 & 84.0 & 77.2 & 77.2 & 59.7 \\
    \hline
    \multirow{3}{*}{XC7Z045} & 1:0 & 41830 & 900 & 160 & 31293 & 144.7 & 129.6 & 143.6 & 91.3 & 89.6 & 108.0 \\
    ~ & 1:1 & 93440 & 900 & 194 & 65699 & 285.5 & 258.1 & 283.7 & 183.2 & 212.5 & 217.2 \\
    ~ & 1:2 (opt.) & 145049 & 900 & 225.5 & 111575 & 359.2 & 326.9 & 390.0 & 318.2 & 369.2 & 340.7 \\
    \bottomrule
\end{tabular}
\end{table*}

The optimal MSQ implementations of CNNs based on ImageNet and previous designs are compared in Table~\ref{tab:comp}, from which it can be observed that our ResNet-18 implementations achieve the highest accuracy and enjoy comparable hardware utilization efficiency represented by GOPS/DSP and GOPS/kLUT with designs in~\cite{guo2017angel,yang2019synetgy}. The work~\cite{wang2018design} acquires higher utilization efficiency but much lower accuracy. 
MobileNet-v2 has the most complicated structure among all these networks, making it difficult to deploy on hardware platforms, but our designs can still achieve high performance, especially in terms of frame rate. We do not find implementations with ResNet-18 and MobileNet-v2 in other work, so we compare it with other CNNs.

Our proposed solution is beneficial over low-precision GPU for the following two reasons: (1) Current low-precision GPU (Tensor-RT solution) relies on 8-bit, while we can go to 4-bit and further assisted by SP2; (2) FPGA solution is dataflow-based and energy-efficient in general \cite{cong2018understanding}. Comparing with a state-of-art energy-efficient GPU (NVIDIA Jetson AGX, power consumption 10-15W) with Tensor-RT support, we use ResNet-18 as example, measured under the same accuracy. Our FPGA solution (XC7Z045) is slightly higher performant (99FPS vs. 78FPS), but more than 3$\times$ higher energy efficiency as the FPGA only consumes around 4W power.

\begin{table*}[htb]
\caption{\textbf{Comparisons of CNNs on ImageNet with previous implementations.}}
\label{tab:comp}
\centering
\begin{tabular}{c|ccc|c|c|cc|cc}
    \toprule
    \multirow{2}{*}{Implementation} & \multicolumn{3}{c|}{VGG} & AlexNet & DiracDeltaNet & \multicolumn{2}{c|}{ResNet-18} & \multicolumn{2}{c}{MobileNet-v2} \\
    ~ & \multicolumn{3}{c|}{\cite{guo2017angel}} & \cite{wang2018design} & \cite{yang2019synetgy} & \multicolumn{2}{c|}{(Our opt.)} & \multicolumn{2}{c}{(Our opt.)} \\
    \hline
    Device & XC7Z045 & XC7Z045 & XC7Z020 & XC7Z045 & XCZU3EG & XC7Z020 & XC7Z045 & XC7Z020 & XC7Z045 \\
    \hline
    Bit-width (W/A) & 16/16 & 8/8 & 8/8 & 8/8 & 1/4 & \multicolumn{2}{c|}{4/4} & \multicolumn{2}{c}{4/4} \\
    \hline
    Top-1 Accuracy & 67.84\% & 67.72\% & 67.62\% & 54.6\% & 68.5\% & \multicolumn{2}{c|}{70.27\%} & \multicolumn{2}{c}{65.64\%} \\
    \hline
    Frequency (MHz) & 150 & 150 & 214 & 200 & 250 & \multicolumn{2}{c|}{100} & \multicolumn{2}{c}{100} \\
    \hline
    LUT & 182616 & 139385 & 29867 & 86262 & 24130 & 28288 & 145049 & 28288 & 145049 \\
    DSP & 780 & 900 & 190 & 808 & 37 & 220 & 900 & 220 & 900 \\
    BRAM36 & 486 & 390.5 & 85.5 & 303 & 170 & 56 & 225.5 & 56 & 225.5 \\
    \hline
    Throughput (GOPS) & 187.8 & 292 & 84.3 & 493 & 47.09 & 77.0 & 359.2 & 71.8 & 326.9 \\
    Frame Rate (FPS) & 6.06 & 9.42 & 2.72 & 340 & 96.5 & 21.3 & 99.1 & 120.7 & 549.3 \\
    GOPS/DSP & 0.241 & 0.324 & 0.444 & 0.610 & 1.273 & 0.350 & 0.391 & 0.326 & 0.363 \\
    GOPS/kLUT & 1.029 & 2.096 & 2.825 & 5.747 & 1.953 & 2.725 & 2.475 & 2.538 & 2.252 \\
    \bottomrule
\end{tabular}
\end{table*}
\section{Related Work}\label{sec:relatedwork}

{This section introduces the DNN weight quantization methods/algorithms for fixed-point and P2 quantization schemes, and discusses DNN weight quantization on FPGA platforms.}

\subsection{DNN  Quantization Methods}\label{sec:related_quantization_methods}

Zhou et al. \cite{zhou2016dorefa} first explored the potential of fixed-point quantization by introducing hyperbolic tangent transformation to weights and activations, with scaling factors to minimize quantization error. 
Choi et al. \cite{choi2018pact} improved this method by adding a parameterized clipping threshold to activations. 
As alternatives for solving the non-differentiable problem, DSQ \cite{gong2019differentiable} developed an evolving training method to gradually approaximate STE.
QIL \cite{jung2019learning} parameterized the quantization interval and trained it with task loss, avoiding access to the original training data. $\mu$L2Q \cite{cheng2019uL2Q} introduced data distribution loss during training to minimize quantization error. LQ-Net \cite{zhang2018lq} and LSQ \cite{esser2019learned} proposed a differentiable method to learn the quantizer for each layer jointly with parameters. Miyashita et al. \cite{DBLP:journals/corr/MiyashitaLM16} replaced fixed-point quantizer with logarithmic representation to exploit bit shift operations to accelerate inference. INQ \cite{DBLP:journals/corr/ZhouYGXC17} splits weights into groups and iteratively quantize the model to low bit-width. Leng et al. \cite{leng2018extremely} employed ADMM training technique to increase the accuracy of extremely low bit-width DNNs. 

In addition to these quantization methods for inference acceleration, Zhu et al. \cite{zhu2020towards} proposed a low-bit training framework for the training acceleration. They used the direction sensitive gradient clipping and the deviation counteractive learning rate scaling to ensure a unified 8-bit (INT8) training with minor accuracy degradation.


\subsection{Weight Quantization in FPGA Implementations}

{Weight quantization has been widely applied to DNN implementations on FPGAs~\cite{nn-accelerator}.
Some work studies fixed-point quantization. The work~\cite{guo2017angel} utilizes greedy solution to determine the radix position of each layer for quantization.~\cite{wang2018design} investigates a hybrid quantization scheme that allows different bit-widths for weights, providing more flexibility.
For Binarized Neural Networks (BNNs), multiplications can be executed with XNOR gates~\cite{nakahara2016memory,nakahara2017fully,guo2018fbna}. A fully binarized neural network accelerator is implemented in~\cite{guo2018fbna} through utilizing odd-even padding to replace the zero padding values.
Another scheme called logarithmic quantization using power of 2 is explored in~\cite{luo2019rna}.
In addition, weight quantization could be employed with a two-stage arithmetic unit for low bit-width CNNs~\cite{jiao2017accelerating}, a fast matrix and Winograd algorithm~\cite{wu2018novel}, a novel CNN architecture for software-hardware codesign~\cite{yang2019synetgy}, a design flow of DNN implementations for more flexible quantization schemes~\cite{zhang2018dnnbuilder}, and an OpenCL-based framework Deep Learning Accelerator (DLA) to accomodate designs with different bit-widths~\cite{colangelo2018exploration}.
In addition, dynamic quantization with bit fusion in \cite{sharma2018bit} improves the bit-level flexibility by matching various bit-widths for different DNN layers.
}

\section{Conclusion}\label{sec:conclusion}

This paper investigates efficient DNN inference engine on FPGA devices through DNN quantization, and proposes the {\em first} solution that applies different quantization schemes for  different rows of the weight matrix. 
We propose a hardware-friendly quantization scheme named {\em SP2} suitable for Gaussian-like weight distribution, in which the multiplication arithmetic can be replaced with logic shifter and adder, thereby enabling highly efficient implementations with the FPGA LUT resources.
In contrast, the fixed-point quantization is suitable for Uniform-like weight distribution and can be implemented efficiently by DSP.
To fully explore the FPGA resources, intra-layer, multi-scheme quantization framework with an ensemble of the SP2 and fixed-point schemes. 
We evaluate our FPGA-centric quantization framework across multiple application domains with various DNNs such as convolutional neural networks (CNN) and recurrent neural networks (RNN).
With optimal SP2/fixed-point ratios on two FPGA devices, i.e., Zynq XC7Z020 and XC7Z045, we achieve performance improvement of $2.1\times-4.1\times$ compared to solely exploiting DSPs for all multiplication operations.

\section*{Acknowledgment}

This work is partly supported by the National Science Foundation CCF-1901378, CCF-1919117, CCF-1919289, CNS-1909172 and DARPA-HR00112090055.

\bibliographystyle{IEEEtran}
\bibliography{refs}

\end{document}